# Automatic Contact-Based 3D Scanning Using Articulated Robotic Arm

Shadman Tajwar Shahid[1,*], Shah Md. Ahasan Siddique[2], Md. Humayun Kabir Bhuiyan[3]

[1,2]Department of Mechanical Engineering, Military Institute of Science and Technology, Dhaka-1216, Bangladesh
[3]Bangladesh Army University of Science and Technology, Khulna, Bangladesh
*Corresponding author: shadmantajwar1997@gmail.com

**Abstract.** This paper presents an open-loop articulated 6-degree-of-freedom (DoF) robotic system for three-dimensional (3D) scanning of objects by contact-based method. A digitizer probe was used to detect contact with the object. Inverse kinematics (IK) was used to determine the joint angles of the robot corresponding to the probe position and orientation, and straight-line trajectory planning was implemented for motion. The system can take single-point measurements and 3D scans of freeform surfaces. Specifying the scanning area's size, position, and density, the system automatically scans the designated volume. The system produces 3D scans in Standard Triangle Language (STL) format, ensuring compatibility with commonly used 3D software. Tests based on ASME B89.4.22 standards were conducted to quantify accuracy and repeatability. The point cloud from the scans was compared to the original 3D model of the object.

## INTRODUCTION

3D scanning plays an important role in industries by enabling documentation and analysis of objects, facilitating design verification, reverse engineering, and modifying products already in the market. There are two primary modes of 3D scanning: contact and non-contact. Contact scanning methods rely on physical interaction with the object, using digitizing probes or styluses to measure its dimensions directly. Non-contact scanning techniques eliminate the need for physical touch, using lasers, cameras, or sound waves to capture data from a distance.

Ultrasonic (US) scanning produces a 2D image of the internals of the object of interest. A 3D image can be created by combining the collection of raw 2D scans in correspondence with their spatial information. US scanning probes, usually handheld, are prone to errors in volume construction due to human afflictions such as hand tremors, fatigue, etc. To overcome this drawback, researchers have employed robotics arms or other types of mechanical manipulators to smoothly move the probe in a pre-defined path [1].

In [2], a 1 DoF linear sliding track fitted with position sensors was used with a handheld probe to produce 3D scans. The system had limited flexibility due to reduced DoF and difficulties maintaining homogenous tissue pressure. In [3], a 6 DoF robotic system was developed that uses a normal-vector-based method on depth camera scans of the object to determine the probe's path for the US scan automatically. Force sensors were used to maintain pressure on the tissue. In [4], a US range finder was used to map the contour of an object's surface using a Cartesian robot, and the experiments obtained errors of up to 2 cm.

In [5], a robotic system for 3D scanning was proposed that detects holes in the model and automatically moves the scanner to an appropriate position to close the holes and get a complete model. A major challenge of using such scanners in conjunction with a robotic arm is the isolated nature of a robotic controller [6]. Integrating the scanner and robotic arm uses a separate PC that receives the robot's end effector spatial position and orientation from the robot controller and combines the transformations to the scanner output.

The coordinate measuring machine (CMM) belongs to the group of contact measuring devices. Gantry-type CMMs are very large, have long installation procedures, and are fixed in place. Most Gantry-type CMMs can measure the dimensions automatically [7]. Simpler and cheaper contact measuring devices are Articulated Arm Coordinate Measuring Machines (AACMM), which are small, medium-sized, and usually portable. The AACMM usually has three links with angular encoders embedded at the joints. The operator manually measures by pressing a button. AACMMs offer a good price-efficiency ratio and are broadly used in the industry. Their main disadvantages are the lack of automation and lesser accuracy [7]. This paper proposes a 6 Degree of Freedom (DoF) robotic arm system called Articulated Robotic Arm Coordinate Measuring Machines (ARACMM) that uses a contact-based method for 3D scanning. Using a robotic arm, the objective is to merge the versatility of an AACMM with the automation capabilities inherent in a gantry-type CMM. Addressing the cost challenge of ultrasonic (US) or fixed-stripe laser scanners, the goal is to develop a cost-effective system by converting any robotic arm into a scanner with the addition of a digitizer probe, essentially a switch. The inverse kinematics (IK) equation was derived, and a straight-line motion planning approach was implemented. Furthermore, the program was designed to produce a 3D file in the Standard Triangle Language (STL) format. Both the control of the robotic arm and the processing of the 3D model were executed within the robot controller.

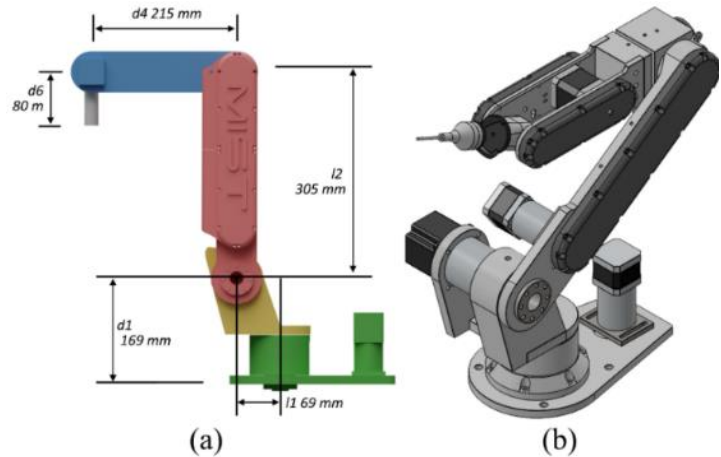

**Figure 1.** (a) Dimensions of robotic arm. (b) Isometric view of robotic arm equipped with digitizer probe.

## METHODS

### System Design

The experimental setup comprises a 6-DoF articulated robotic arm with a 1 kg payload capacity and 600 mm reach, a digitizing probe, an Arduino Mega 2560 microcontroller, and a personal computer (PC). The 6-Dof articulated robotic arm was adapted from Annin Robotic's open-source AR3 model. Stepper motors drive the arm and use planetary gears for speed reduction. All the structural elements other than the shafts were 3D printed. The dimensions of the robot are illustrated in Figure 1(a). The 5-axis digitizer probe is based on the Renishaw company touch probe, which uses the kinematic coupling mechanism [8]. Kinematic couplings restrain 6 degrees of freedom between two components, the body and stylus, and maintain high repeatability.

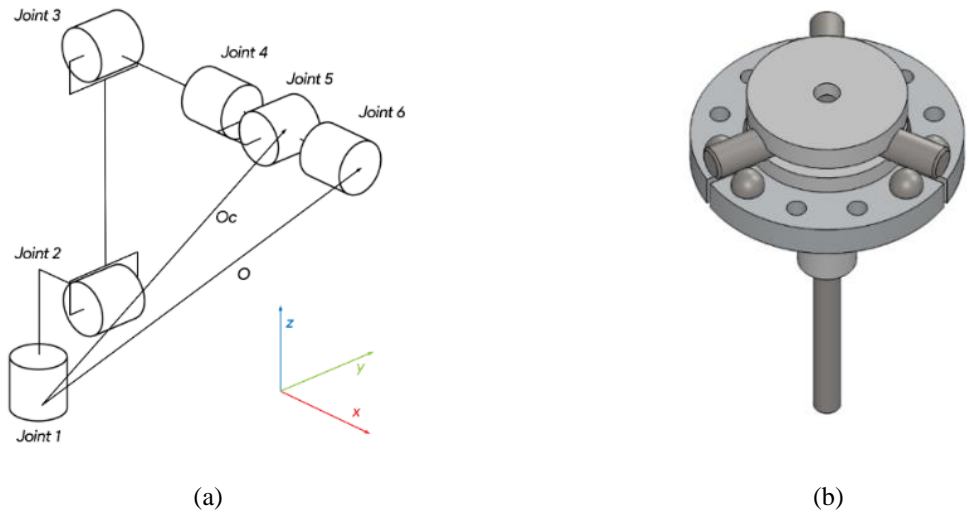

**Figure 2.** (a) Kinematic decoupling. (b) Construction of digitizer probe.

# Inverse Kinematics

The position and orientation of the robotic arm's end effector in 3D space are given as a 4X4 matrix representation of the end frame, where the last column is the body's position, and the first three columns are the values of Euler angles.

$$T_6^0 = \begin{vmatrix} n_x & o_x & a_x & p_x \\ n_y & o_y & a_y & p_y \\ n_z & o_z & a_z & p_z \\ 0 & 0 & 0 & 1 \end{vmatrix} = \begin{vmatrix} r_{11} & r_{12} & r_{13} & p_x \\ r_{21} & r_{22} & r_{23} & p_y \\ r_{31} & r_{32} & r_{33} & p_z \\ 0 & 0 & 0 & 1 \end{vmatrix} \qquad (1)$$

Using Inverse Kinematics (IK), the joint angles that the robotic arm needs to make to reach the given target are derived. By using Piper's approach [10], which divides the robotic arm into a spherical wrist and 3 DoF robotic arm, the IK of the robotic arm is analyzed. The first step is to find the position of the end of the 3 DoF arm, for any given orientation and position of the end-effector as illustrated in Figure 2(a).

$$\begin{vmatrix} x_c \\ y_c \\ z_c \end{vmatrix} = \begin{vmatrix} p_x - d_6 r_{13} \\ p_y - d_6 r_{23} \\ p_z - d_6 r_{33} \end{vmatrix} \qquad (2)$$

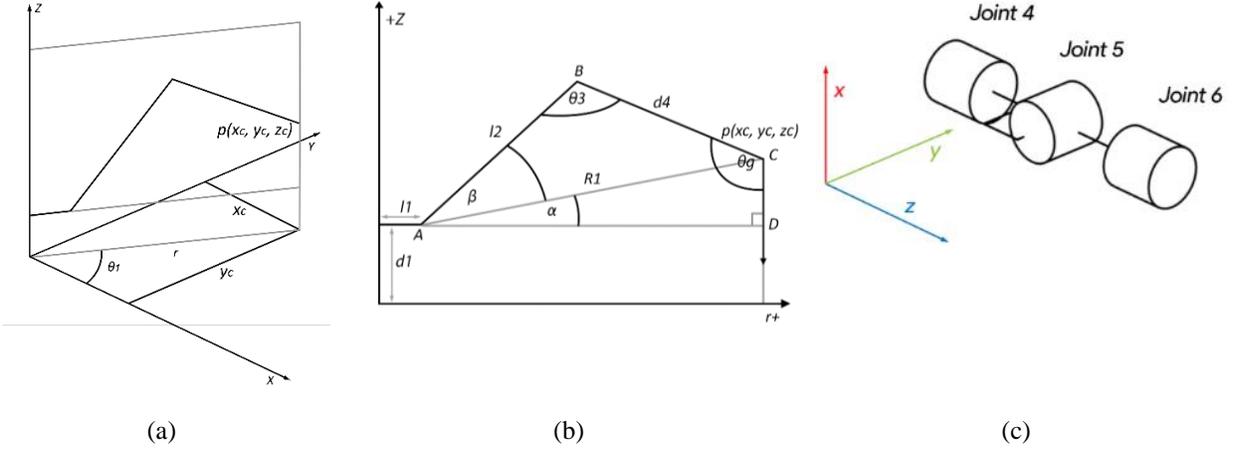

(a)  (b)  (c)

**Figure 3.** (a) Projection of arm on x, y plane. (b) Projection of arm on z, r plane. (c) Spherical joint.

For finding the first three joint angles $\theta_1$, $\theta_2$, and $\theta_3$, joint 2 and joint 3 are projected on the x-y plane as illustrated in Figure 3(a). The projection gives the triangle from which the angle of the joint 1, $\theta_1$ is derived.

$$\theta_1 = \tan^{-1} \frac{y_c}{x_c} \qquad (3)$$

Another projection of joint 2 and joint 3 on the r-z plane as illustrated in Figure 3(b), is used to find the other two angles, $\theta_2$, and $\theta_3$.

$$z = z_c - d_1 \qquad (4)$$

$$R = \sqrt{x_c^2 + y_c^2} - l_1 \qquad (5)$$

$$R_1 = \sqrt{R^2 + z^2} \qquad (6)$$

$$\alpha = \tan^{-1} \frac{z}{R} \qquad (7)$$

Applying the law of cosine in triangle ABC,

$$\beta = \cos^{-1}\frac{l_2^2 + R_1^2 - d_4^2}{2l_2 R_1} \tag{8}$$

$$\theta_2 = \alpha + \beta \tag{9}$$

Again, applying the law of cosine in triangle ABC,

$$\theta_3 = \cos^{-1}\frac{l_2^2 + d_4^2 - R_1^2}{2l_2 d_4} \tag{10}$$

The orientation of the last three joints can be derived from the robot's overall orientation.

$$\theta_5 = \cos^{-1}(r33) \tag{11}$$

$$\theta_4 = \tan^{-1}\frac{r23}{r13}$$

$$\theta_6 = \tan^{-1}\frac{r32}{r31}$$

## Scanning Pattern and Processing

The robot scans the object in a grid pattern for 3D scanning of freeform surfaces. The work envelope is set by defining the coordinates of a corner that will be the first point, the number of rows and columns, and the distance between these points. The motion is illustrated in Figure 4. The Standard Triangle Format (STL) was selected as the output of the 3D scan. An STL file describes an object's surface using groups of triangles defined by the Cartesian coordinates of their three vertexes and the corresponding plane's normal vector.

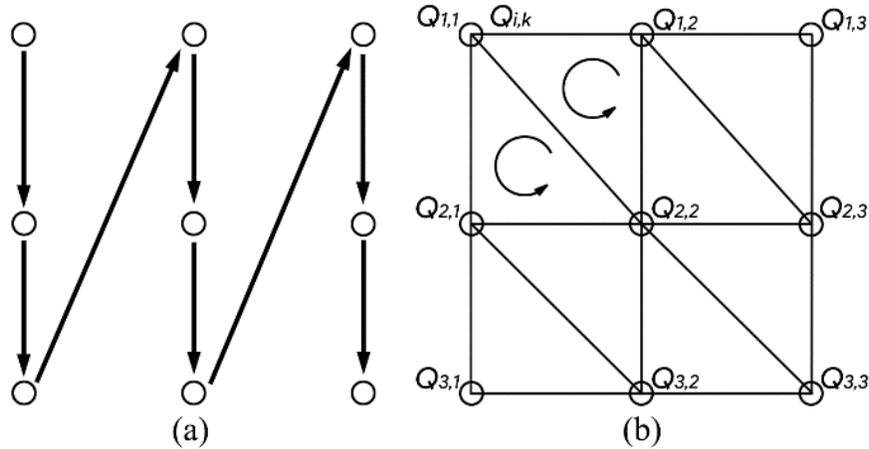

**Figure 4.** (a) Scanning pattern. (b) Triangle generation.

As the digitizer moves along the first column, no triangle is produced. Instead, the height values are stored in an array. $Q_{i,k}$ is an arbitrary point in the grid. When the arms move to the 2nd column after reaching the total rows and the 2nd row of that column, that is $i > 1$, $k > 1$, are two triangles produced in the format illustrated in Figure 7(b). The sequence of vertex coordinates of the two triangles is given below,

$$Q_{i,k} \rightarrow Q_{i-1,k} \rightarrow Q_{i-1,k-1}$$

$$Q_{i,k} \rightarrow Q_{i-1,k-1} \rightarrow Q_{i,k-1}$$

# ANALYSIS

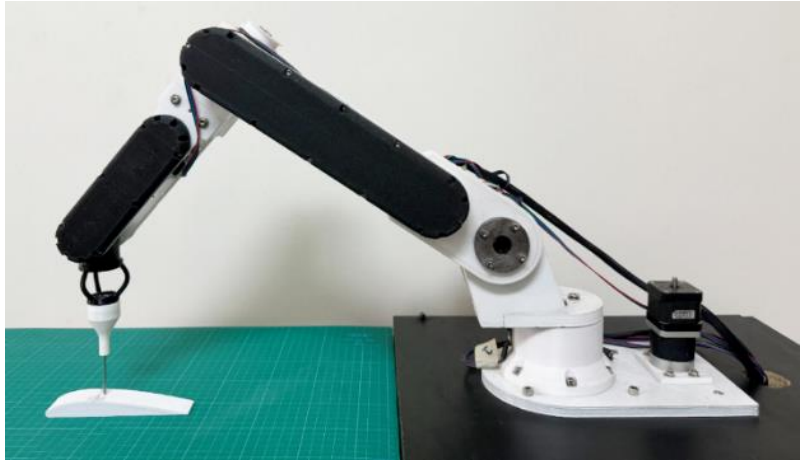

**Figure 5.** Experimental setup for scanning airfoil

## Similarity Test of Scanned Surface

A widely used mathematical metric to quantify how similar two 3D models are in terms of their shape is the Chamfer Distance (CD). CD calculates the mean distance from each point in one set to the nearest point in the other set, and vice versa. P = {$p_1, p_2,…, p_m$} and Q = {$q_1, q_2,…, q_n$} be the two point clouds, one scanned from the experiment and another from the object's original STL file respectively. The equation to calculate the chamfer distance from point cloud P to Q is as follows-

$$CD(P,Q) = \frac{1}{m}\sum_{i=1}^{m} \min_{j=1} \left\| p_i - q_j \right\| + \frac{1}{n}\sum_{j=1}^{n} \min_{i=1} \left\| q_j - p_i \right\| \qquad (12)$$

$||p_i - q_j||$ represents the Euclidean distance between point $p_i$ in set P and its nearest neighbor in set Q and vice versa for $||q_j - p_i||$.

The test object is a 3D-printed wing with a NACA 6409 profile of chord length of 140 mm and a length of 150 mm. The distance between adjacent points along the row and column directions is set to 6 mm. A total of 20 readings are taken along the column and 25 along the row, resulting in a total of 500 points captured per scan, 18 points per square inch. Figures 7 and 8 show the scanned output and a cross-section respectively.

**TABLE 1:** CHAMFER DISTANCE (CD) VALUES

| No. | CD (mm) |
|---|---|
| 1 | 0.822 |
| 2 | 0.887 |
| 3 | 0.810 |
| 4 | 0.781 |
| 5 | 0.792 |
| 6 | 0.803 |
| 7 | 0.861 |
| 8 | 0.823 |
| 9 | 0.862 |
| 10 | 0.824 |
| Average | 0.823 |

# Point Accuracy and Repeatability Test

The ASME B89.4.22 standard quantifies accuracy and repeatability and outlines the methods for evaluating AACMMs [16]. Although AACMMs are not motorized and are manually operated devices, this standard provides a relevant framework for evaluating the robotic system's single-point measurement. The standard has advised three accuracy tests: A, B, and C.
Test A- Sphere test
Test B- Single-point test
Test C- Volumetric Performance test

Test A involves measuring nine coordinates distributed equally on the surface of a sphere with a nominal diameter ranging from 10 to 50 mm. These coordinates are used to calculate the sphere's radius, and the probe error is determined. The positions of the probe points on the sphere's surface are presented in Figure 6.

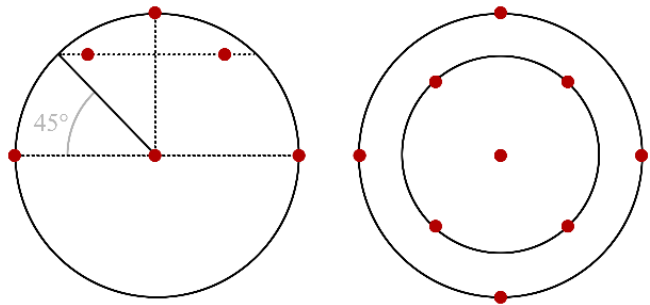

**Figure 6.** Distribution of probe points on the sphere.

Table 2 presents the measured point accuracy measured by Test A.

**TABLE 2:** RESULT OF ACCURACY TEST- TEST A

| Latitude (degrees) | Longitude (degrees) | Accuracy, Diameter difference, $\Delta d$ (mm) |
|---|---|---|
| 0 | 0 | 0.0245 |
|  | 90 | 0.0311 |
|  | 180 | 0.0286 |
|  | -90 | 0.0320 |
| 45 | 0 | 0.0467 |
|  | 90 | 0.0481 |
|  | -180 | 0.0500 |
|  | 90 | 0.0460 |
| 90 | 0 | 0.0213 |
| Average |  | 0.03618 |

Test B involves measuring a single point located at varying distances from the base of the ARACMM. The test assesses the robot's capacity to consistently obtain coordinates when measuring the same point repeatedly, thereby evaluating its repeatability. Table 3 presents the measured point repeatability measured by Test B.

**TABLE 3:** RESULT OF REPEATABILITY TEST- TEST B

| Point distance (mm) | Point repeatability (mm) |
|---|---|
| 120 | +/- 0.0387 |
| 300 | +/- 0.0544 |
| 500 | +/- 0.0712 |

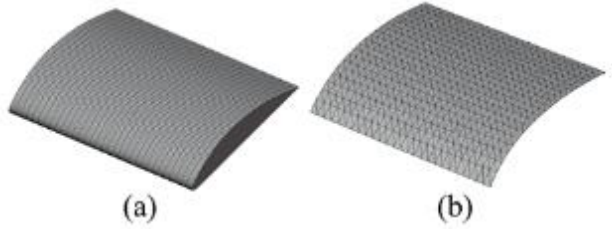

**Figure 7.** (a) STL model of airfoil. (b) Model of the scanned airfoil.

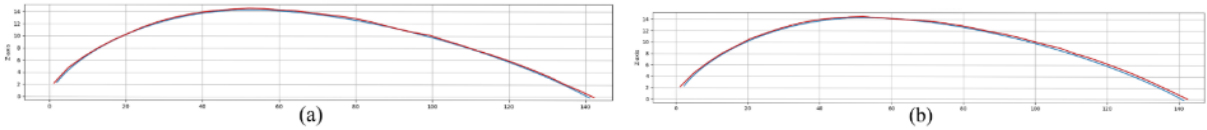

**Figure 8.** (a) Cross-section of the first column. (b) Cross-section of the last column.

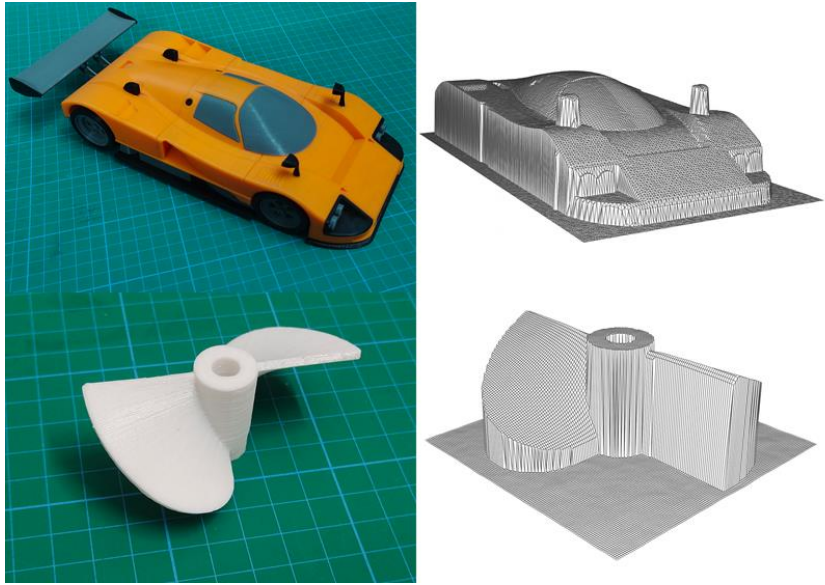

**Figure 9.** Objects (left) and the 3D scanned model (right).

## RESULTS AND DISCUSSIONS

The system performance for 3D scanning was evaluated for two categories: scanning freeform objects and Point-to-Point Measurement. The CD values for freeform scanning are reported in the table. The ARACMM shows higher accuracy and precision than machines in the same price range [16].

Figure 7 shows the 2D cross-sections along the first and last column of the wing. Initially, the scanned profile closely follows the contour of the wing section. However, a small deviation is noticed in the last column towards the end of the scan. The deviation is because of the robot's sudden stop upon contact with the surface. The robot's inertia prevents it from halting abruptly, leading to collisions with the rigid surface of the object. As a result, the joint physically deviates from the expected angle, and as the number of point increases, the deviation compounds and increases the system's inaccuracy. Since the robot is an open-loop system, it cannot compensate for these deviations.

Figure 9 presents the 3D scanned model of more complex objects. However, CD calculations were not performed due to the large number of points measured and the time required. The scanned models show similarity to the actual objects, capturing details and maintaining good accuracy. The system cannot scan the overhanging section of the airfoil's leading edge or the underside of objects. This limitation is due to the probe's inability to reach the bottom surface because of its vertical approach toward the object.

## CONCLUSION

In this paper, an open-loop robotic system was developed to produce 3D scans of objects automatically. A 6-DoF articulated robotic arm moved the digitizer probe in a grid pattern and recorded the contact coordinates. Inverse kinematics and motion equations were derived and implemented to drive the robot. In the end, an STL file and point cloud of the object were produced from the coordinates. Furthermore, accuracy and repeatability tests were conducted, aligning with methods used for AACMMs. The measurements show that it can scan with good accuracy and repeatability. The main advantage of this system is its versatility. Any robotic arm with more than 3 DoFs can be repurposed into a 3D scanner with the addition of an inexpensive digitizer probe. The system is significantly smaller than Cartesian CMMs with equivalent scan volumes and is easily portable. Moreover, the ability to produce the scan as an STL file makes it compatible with industrial use.

The majority of the robot was 3D printed because cost reduction was prioritized. While this effectively lowered expenses, it also introduced problems such as part fatigue, bending, and backlash, reducing the system's accuracy. The fabrication of structural parts from Aluminium would provide a solution. These enhancements are targeted for implementation in the next iteration of our research.